\documentclass{article}
\usepackage{ijcai18}

\usepackage{times}
\usepackage{xcolor}
\usepackage{soul}
\usepackage[utf8]{inputenc}
\usepackage[small]{caption}
\usepackage{amsfonts}
\usepackage{amsmath}
\usepackage{subfigure}
\usepackage{graphicx}
\usepackage{booktabs}
\usepackage{tabularx}
\usepackage{siunitx}
\usepackage{multirow}
\usepackage{tikz}
\usepackage{latexsym}
\usepackage{newtxmath}

\title{Motif-based Convolutional Neural Network on Graphs}

\renewcommand*{\thefootnote}{\fnsymbol{footnote}}
\author{Aravind Sankar\footnotemark[2]\footnotemark[1], Xinyang Zhang\footnotemark[2]\footnotemark[1], Kevin Chen-Chuan Chang\footnotemark[2] \\
\footnotemark[2]Department of Computer Science, University of Illinois at Urbana-Champaign, USA \\
\{asankar3, kcchang, xz43\}@illinois.edu \\
}

\begin{document}

\maketitle

\begin{abstract}
\small
This paper introduces a generalization of Convolutional Neural Networks (CNNs) 
to graphs with irregular linkage structures, especially heterogeneous graphs with typed nodes and schemas.
We propose a novel spatial convolution operation to model the key properties of local connectivity and translation invariance, using high-order connection patterns or \textit{motifs}.
We develop a novel deep architecture \textit{Motif-CNN} that employs an attention model to combine the features extracted from multiple patterns, thus effectively capturing high-order structural and feature information. 
Our experiments on semi-supervised node classification on real-world social networks and multiple representative heterogeneous graph datasets indicate significant gains of 6-21\% over existing graph CNNs and other state-of-the-art techniques.
An updated, published version of this paper can be found here~\cite{metagnn}
\normalsize
\end{abstract}

\section{Introduction}
\label{sec:intro}
\noindent
CNNs~\cite{cnn} have achieved immense success in various machine learning tasks where the underlying data representation is grid-structured, such as image classification, object detection etc.~\cite{deep}. Their efficacy has inspired the study of CNN paradigm in non-euclidean domains such as Natural Language~\cite{sentence_cnn}, Graphs~\cite{iclr14} and Manifolds~\cite{mesh}.

Two fundamental properties utilized in the design of CNNs are \textit{local connectivity} and \textit{translation invariance}. The neurons in a CNN model \textbf{spatial locality} by extracting features from small localized regions called \textit{receptive fields} and achieve \textbf{weight-sharing}
across spatial locations thus capturing translation invariance. 
This enables them to significantly reduce the number of parameters without sacrificing the ability to extract informative patterns.

In many scenarios, we encounter data lying in irregular domains that are naturally represented as graphs encoding interactions between various real-world entities. 
For instance, an academic citation network (such as DBLP) is composed of 
multiple types of nodes, viz. authors, papers and venues inter-connected in different relationships. 

Direct generalizations of CNNs to graphs is non-trivial since real-world graphs do not share the same locality patterns as in grid-structured data. For example, convolutional filter design in images directly follows from the structure of the data manifold where every pixel has eight neighbors with precise spatial locations. Graphs however possess irregular neighborhood structures with variable neighbors per node, rendering standard notions of connectivity and translation inapplicable. 
Besides, most modern graphs are heterogeneous, comprising nodes of several types with diverse feature sets, that entails fine-grained modeling of interactions between different types subject to the schema and domain semantics.
 
The two properties identified earlier for traditional CNNs are quite relevant and in fact necessary for graphs.
The goal of graph convolution is to model a \textit{target} node of interest by extracting features from semantically relevant \textit{context} nodes, which is precisely captured by the property of local connectivity. 
Consider a heterogeneous graph where not every pair of node types can 
connect directly (e.g., in DBLP, an author does not link to another author; they connect through a paper). For an author node, her co-authors and published venues are examples of high-order context nodes 
that provide relevant features.
In addition, the role of various context nodes must be appropriately differentiated to accurately capture different semantics. Thus, we seek to realize the two key properties via appropriate definitions of \textit{spatial locality} and \textit{weight-sharing}, which leads to two key requirements:\\
\textbf{R1} Spatial locality to identify the receptive field around a node, subject to the diversity of node types, semantics and heterogeneous interactions.  \\
\textbf{R2}  Weight-sharing scheme to assign nodes the same weight in different locations of the receptive field if and only if their semantic roles (position in the receptive field) are identical.

Though the requirements outlined above seem natural, existing graph CNNs fall short of addressing them satisfactorily.
Recent work on graph CNNs fall into two categories: spectral and spatial.
Spectral techniques focus on defining convolution using an element-wise product in the Fourier domain, while spatial techniques define convolution through weight-sharing among local neighborhoods of the graph.

Spectral CNNs employ the analogy between classic Fourier transforms and projection onto the eigenbasis of the graph Laplacian to define spectral filtering~\cite{iclr14,nips16}.
However, the  spectral filters are functions of the Laplacian eigenbasis 
and thus incapable of application on another graph with a different structure.

On the other hand, spatial CNNs define spatial locality (R1) based on adhoc definitions of neighborhood proximity that operate on the immediate neighbors of each node~\cite{iclr17,dcnn,graph-cnn}. While immediate neighbors are adequate to model regular grid-structured data, the notion of closeness on graphs is application dependent. As illustrated earlier, we require a high-order notion of locality that is not limited to immediate neighbors to extract relevant features for an author in DBLP. Furthermore, existing spatial CNNs are type-agnostic, thus failing to capture semantic dependencies between nodes of different types. 
Thus, an appropriate definition of spatial locality, subject to graph heterogeneity and semantics is the \textbf{first key challenge} in designing an effective graph CNN.

Weight-sharing schemes (R2) proposed by previous spatial CNNs fail to distinguish semantic roles of nodes in the receptive field. 
Existing schemes work by:
hop distance from a node~\cite{dcnn}, linearizing the neighborhood through a canonical ordering~\cite{icml16}, or aggregation of immediate neighborhood~\cite{iclr17,graph-cnn,graphsage}.
However, they operate under the assumption of a homogeneous neighborhood structure and fail to account for varying semantics of different context nodes, especially in expressive heterogeneous graphs. 
We identify the definition of a weight-sharing scheme that clearly delineates the semantic roles of nodes in the receptive field 
as our \textbf{second key challenge}.  \\[2pt]
To address the above challenges, we propose \textit{motifs} to model the receptive field (the central notion of a CNN) around a target node of interest. Motifs, also known as high-order structures, are fundamental building blocks of complex networks,~\cite{complex}, that describe small subgraph patterns with specific connections among different node types. We identify two key insights captured by the use of motifs: 

1. \textbf{High-order locality:} Unlike adhoc definitions of local neighborhood, motifs specify the context nodes \textit{relevant} to a target node of interest linked via certain patterns, thus providing a principled framework capturing high-order locality, such as an author (target) connecting to another author (context) through a paper they coauthored.

2. \textbf{Precise semantic role:} Motifs enable accurate discrimination of semantic roles of various nodes in the receptive field based on their types and structural linkage patterns, such as distinguishing the roles of a co-author (context) and publication venue (context) in characterizing a target author. 

Conceptually, motifs are similar to metagraphs~\cite{metagraph} that have been successfully used to model semantic proximity in heterogeneous graphs, and hence can be described through domain knowledge.
We use this as our basis to develop a novel graph CNN architecture \textit{Motif-CNN}.
We summarize the main contributions of our paper below:

1. We use motifs to define the receptive field around a target node of interest modeling the key aspects of local connectivity and translation invariance, thus
capturing high-order semantics in homogeneous and heterogeneous graphs alike.

2. We present a novel motif-based convolution operation that delineates the semantic roles of various nodes in the receptive field and extracts features across motif instances in the local neighborhood. To the best of our knowledge, we are the first to explore motifs for graph convolution.

3. We propose a novel deep architecture \textit{Motif-CNN} that uses an attention mechanism to effectively integrate the features extracted from multiple motifs.
We believe that \textit{Motif-CNN} is the first neural network designed for semi-supervised learning (SSL) that generalizes to heterogeneous graphs.

4. Our experimental results on real-world social networks and multiple heterogeneous graphs demonstrate the effectiveness of our model in leveraging high-order features. 

\renewcommand*{\thefootnote}{\arabic{footnote}}
\setcounter{footnote}{0}

\section{Motif-based Convolutional Neural Network}
\label{sec:gcnn}
We first introduce notations used throughout the paper. 
\subsection{Preliminaries}
\label{sec:prelim}
A graph is defined as $G = (V,E)$ with node type mapping  $l: V \mapsto \mathcal{L}$
where $V = \{ v_1, ..., v_N\}$ is the set of
$N$ nodes, $E$ is the set of edges and $\mathcal{L}$ is the node type set.
We exclude link types for the sake of brevity. 
The nodes are collectively described by a feature matrix $X \in \mathbb{R}^{N \times D}$ where $D$ is the number of features.
Since different node types in a heterogeneous graph may not share the same feature space, we concatenate the features of all types (with zero padding) to get a joint representation of $D$ features.
\subsection{Properties of Convolution}
\begin{figure}[t]
\vspace{-3pt}
\subfigure{
        \centering
        \includegraphics[width=0.55\linewidth]{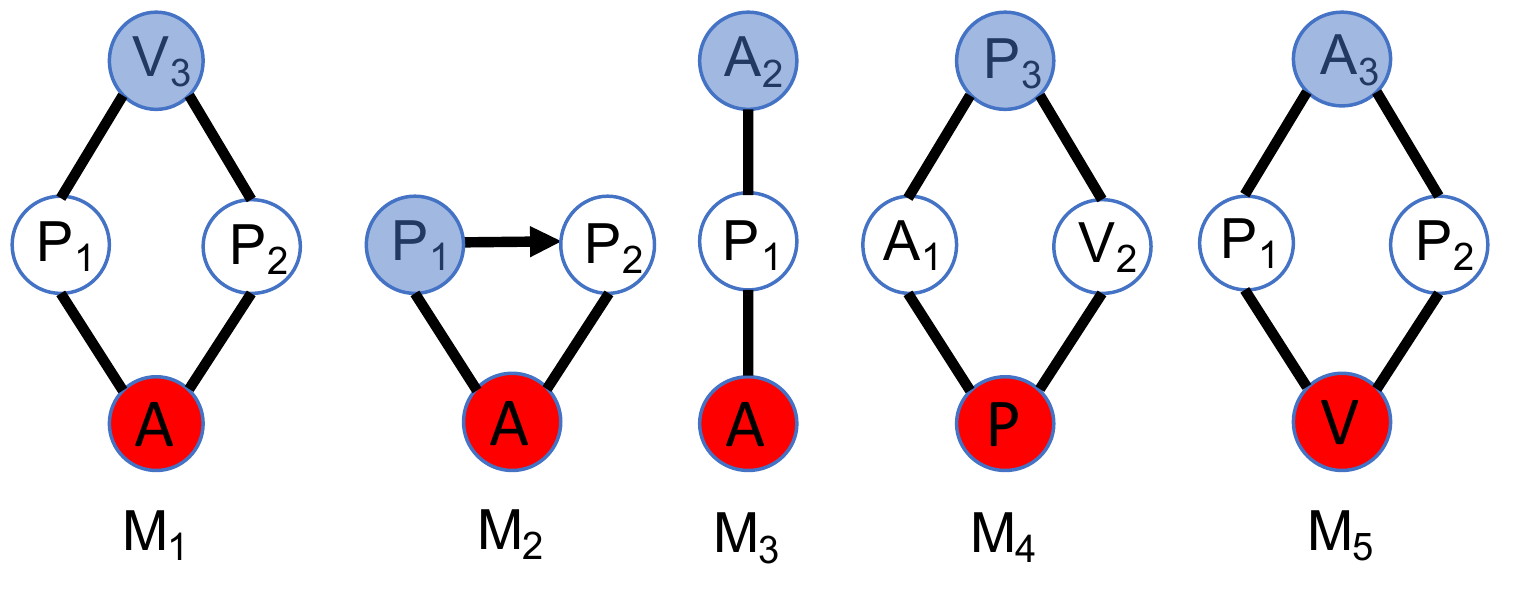}
        \label{fig:motif}}
        \hspace{0.02\linewidth}
\subfigure{
        \centering
        \includegraphics[width=0.35\linewidth]{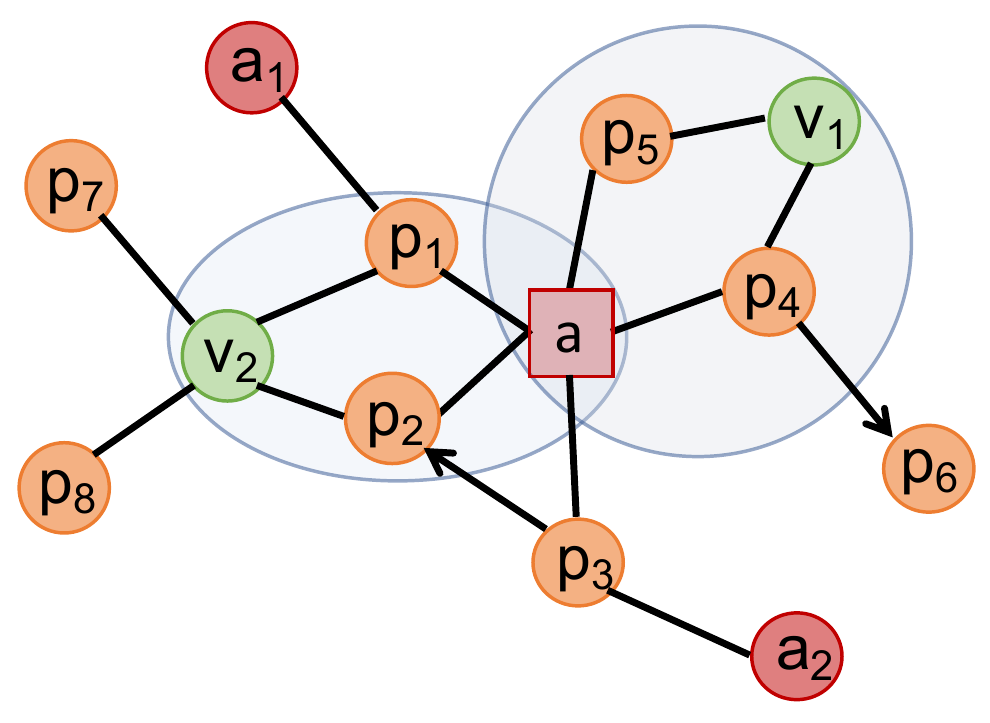}
        \label{fig:conv}}
                \vspace{-13pt}
        \caption{(a) Sample Motifs in DBLP - Types: Author (A), Paper (P) and Venue (V) with target in red and context in blue (b) Example graph with instances of $M_1$ for target $a$.}
        \vspace{-16pt}
\end{figure}
In this section, we revisit the key properties of a CNN to motivate and establish the foundation of our model.

A conventional CNN models the locality of a pixel (receptive field) using a `square-grid' of fixed size and extracts translationally invariant features by scanning a filter across the grid-structured input. In graphs, the goal is to characterize a specific node of interest, \textit{target} node, through features of semantically relevant neighboring nodes, \textit{context} nodes. \\
\textbf{Local Connectivity:} 
Let us revisit, e.g., DBLP, with the goal of predicting the research area of an author. Though an author is linked only to papers, the venues and citations of her published papers provide strong cues in identifying her research area.
It is thus necessary to look beyond the immediate neighbors to model local connectivity through high-order structures or \textit{motifs} that encode specific linkage patterns.

Conventionally, a motif (or metagraph) has been defined as a pattern of edges among different node types~\cite{science,metagraph}.
To model locality specific to a node type, we use a motif to characterize the interaction of a target type with a (possibly) different context type through semantically relevant patterns of connections. 
In Fig.~\ref{fig:motif}, motif $M_1$ is a pattern that describes the interaction of a \textit{target} node $A$ with \textit{context} node $V_3$ through \textit{auxiliary} paper nodes $P_1$ and $P_2$, i.e., high-order structure $M_1$ indicates that $P_1$, $P_2$ and $V_3$ provide relevant features to characterize $A$. 
We use lower-case letters (e.g., $a$) to denote nodes in $G$ and upper-case letters (e.g., $A$) for node types. In Fig.~\ref{fig:motif}, the subscript under a node gives the node index, e.g., $P_1$ and $P_2$ are two different nodes of type $P$. An author node $a$ is now characterized via different instances of motif $M_1$ in $G$ with $a$ as target (illustrated in
Fig.~\ref{fig:conv}), i.e., $a$ is locally connected with the context nodes $v_1$ and $v_2$ in the two marked instances.\\
\textbf{Translation Invariance:} 
Consider Fig.~\ref{fig:conv} where $a_1$ and $a_2$ are linked to target node $a$ via motif $M_3$. They represent co-authors who are expected to share a similar relationship with $a$.
Generalizing this idea, we posit that nodes linked via certain patterns of connections share the same \textit{semantic roles} relative to the target node. Since a motif contains one context node and
multiple auxiliary nodes, it is essential to delineate their semantic roles
relative to the target node to accurately model weight-sharing.
In motif $M_2$ (Fig.~\ref{fig:motif}), it is evident that the roles of $P_1$ and $P_2$ relative to target $A$ are different, while the roles of $P_1$ and $P_2$ are indistinguishable in $M_1$.
Thus, two nodes in the receptive field share the same semantic roles in motif $M$ if and only if they are of the same type and are structurally symmetric relative to the target node. 

We formalize this by using a motif to define a receptive field specific to a node type. A motif $M$ is defined as a subgraph composed of a designated \textit{target} node $t_M$, \textit{context} node $c_M$ and
 \textit{auxiliary} nodes $B_M$, with $K_M$ unique \textit{semantic} roles. \\
\textbf{Definition 1:} \textit{A \textbf{motif} $M$ with \textbf{target} node $t_M$, \textbf{context} node $c_M$ and auxiliary nodes $B_M$ is defined as $M = (V_M, E_M, B_M, t_M, c_M, \phi_M)$ with node type mapping $l_M: V_M \mapsto \mathcal{L}$ where $V_M$ is the set of all nodes with $t_M, c_M \in V_M$  and $B_M \subseteq V_M - \{t_M,c_M\}$, $E_M$ is the set of edges, $\phi_M: V_M - \{t_M\} \mapsto \{1, \dots, K_M\}$ is a role mapping function that returns the semantic role of a node and $ \forall x \in V_M, l_M(x) \in \mathcal{L}$.}

Since the semantic roles of various nodes w.r.t. target node $t_M$ in motif $M$  
can be easily deduced from the structure and types of nodes in $M$, we include it in the definition of a motif. The number of unique roles $K_M$ is at most 1+$|B_M|$.

We define an instance $S_u$ of motif $M$ with target $u$ in $G$ as a subgraph induced by $M$ with $u$ as the target node.\\ 
\textbf{Definition 2:} \textit{An \textbf{instance} $S_u = (V_S, E_S) $ of motif $M$ with target node $u$ on $G$ is a subgraph of $G$ where $V_S \subseteq V$ and $E_S \subseteq E$, such that there exists a bijection  $ \psi_S : V_S \mapsto V_M $ satisfying (i) $u \in V_S, \; \psi_S(u) = t_M $ (ii) $\forall x \in V_S, \; l(x) = l_M(\psi_S(x))$ and (iii)  $\forall x,y \in V_S, \; (x,y) \in E_S$ if $ (\psi_S(x), \psi_S(y)) \in E_M$.}

The receptive field around a target node $t_M$ is defined by the context node $c_M$ and auxiliary nodes $B_M$ in motif $M$. 
We define a tensor to model role-specific motif connectivity.\\
\textbf{Motif-adjacency Tensor:} We define $\mathcal{A}^M$, a tensor of $K_M$ matrices to 
 encode the occurrences of nodes in each unique semantic role $k$ over all instances of $M$ in the graph $G$. $\mathcal{A}^M_{kij}$ is the number of times node $v_j$ appeared in an instance of $M$ in role $k$ with $v_i$ as target. Formally, 
\scriptsize
\begin{align*}
\mathcal{A}^M_{kij} = \sum\limits_{S_{v_i} \in I^M_{v_i}} I \Big( \phi (V_S, E_S, v_i, v_j) =k \Big)
\vspace{-10pt}
\end{align*}

\normalsize
where $I(\cdot)$ is the indicator function. We define a diagonal matrix $\mathcal{D}^M \in \mathbb{R}^{N \times N}$ to store the number of motif instances at each node (as target), i.e. $\mathcal{D}^{M}_{ii} = |I^M_{v_i}| = L_i \; \forall 1 \leq i \leq N$. 

For any graph-based learning task, it is necessary to use a combination of multiple \textit{relevant} motifs to achieve good performance. A single motif 
describes a specific semantic connection pattern among different node types that is permissible by the schema. 
Unlike conventional filters that extract features around each pixel in an image, a motif is useful only for certain nodes in the graph, e.g., a triangular motif can model the context around densely connected nodes but would fail for nodes with only one neighbor and not all node types can be covered by a single motif in heterogeneous graphs. Furthermore, the underlying task at hand may require multiple semantic patterns for optimal performance. Thus, we employ multiple domain-specific \textit{relevant} motifs as input.

We consider a motif \textit{relevant} if the context node can provide useful features to model the target node. Let us consider, e.g., the interaction between a pair of author nodes in DBLP. Since the schema does not permit a direct $A\mbox{-}A$ link, $M_3$ is an example of a motif that describes a co-authorship relation. Since such relevance depends on the specific domains (e.g., authors and publications) and applications (e.g., classifying authors' areas), our framework assumes a set of motifs as input, which are specified by domain experts to capture the relevance between different types of nodes.In our experiments, we explore all \textit{relevant} motifs of upto 3 nodes.\\
\textbf{Problem Definition:}
Given a set of $U$ motifs $U_M = \{ M_1, \dots M_U\}$ as input along with their respective motif-adjacency tensors, our goal is to learn a prediction model on the nodes of $G$ given task-specific supervision. In this paper, we focus on the application of semi-supervised node classification where we are given a set $Y_{{}_L}$ of nodes with labels. 

\begin{table}[t]
\centering
\scriptsize
\begin{tabular}{|c|c|l|}
\hline
& Dimensions & Description \\ \hline
$N$ & Scalar & Number of  nodes \\ \hline
$D$ & Scalar & Number of input features \\ \hline
$F$ & Scalar & Number of filters learnt per motif \\ \hline
$K_M$ & Scalar & Number of  unique semantic roles in motif $M$ \\ \hline
$U$ & Scalar & Number of motifs/Conv. units in one Layer  \\ \hline
$X$ & $N \times D$ & Input Feature Matrix \\ \hline
$H^j_l$ & $ N \times F$ & Activations at Conv unit $j$ (motif $M_j$) in layer $l$ \\ \hline
$\mathcal{A}^M$ & $K_M \times N \times N $ & Motif-Adjacency tensor for motif $M$ \\ \hline
$W^{M}$ & $(K_M+1) \times D \times F$ & Filter weight tensor for motif $M$ \\ \hline
\end{tabular}
\vspace{-8pt}
\caption{Notations}
\normalsize
\vspace{-15pt}
\end{table}

\subsection{Motif-based Convolution}
\label{sec:conv_unit}
In this section, we propose a novel motif-based spatial convolution operation to extract local features for a specific node type, capturing the above described properties. 

Specifically, we are
given motif $M$ with target type $T =  l(t_M)$ and a target node $v_i \in V$ with $l(v_i) = T$ as input.
For ease of explanation, we restrict our initial definitions to a single feature ($D = 1$) and single filter per motif ($F=1$).\\
\textbf{Motif Filter:} A motif filter (on $M$) is defined by a weight $w_0$ for target $t_M$ and a weight vector $\mathbf{w} \in \mathbb{R}^{K_M}$ for the $K_M$ roles, i.e., each weight  in $\mathbf{w}$ differentiates the semantic roles of context and auxiliary nodes in the receptive field. \\
\textbf{Convolutional Unit}: 
To define convolution at node $v_i$, the features of all nodes locally connected through motif $M$ are weighted according to their semantic roles and normalized by the diagonal matrix that reduces the bias introduced by highly connected nodes, giving rise to:
\scriptsize
\begin{equation}
 h^M (v_i) =  \sigma \Bigg(w_0  x_{i} +   \frac{1}{\mathcal{D}^M_{ii}} \sum\limits_{j=1}^{N}  \sum\limits_{k=1}^{K_M}w_k   \mathcal{A}^M_{kij} x_{j} \Bigg)
 \label{eqn:conv_partial}
 \vspace{-5pt}
\end{equation}
\normalsize
where $x_i$ and $x_j$ refer to the features of nodes $v_i$ and $v_j$ respectively,  $h^M(v_i)$ is the output of convolution at node $v_i$ and $\sigma(\cdot)$ is an activation function, such as  $ReLU(\cdot) = \max(0, \cdot)$. We term this a \textit{convolutional unit} (Conv. Unit) for motif $M$.

Thus, 
weight-sharing is achieved by assigning the same weight to context and auxiliary nodes sharing the same semantic role relative to $v_i$ across all instances of $M$. 

On generalizing Eqn.~\ref{eqn:conv_partial} for input $X \in \mathbb{R}^{N \times D}$ with $N$ nodes, $D$ features and $F$ filters per motif, we get:
\scriptsize
\begin{equation}
H^M = \sigma \Bigg( X W_0^M + (\mathcal{D}^{M})^{-1} \sum\limits_{k=1}^{K_M}  \mathcal{A}^M_{k} X W^M_k \Bigg)
\label{eqn:conv_full}
\vspace{-5pt}
\end{equation}
\normalsize
where $W^M$ is a tensor of filter parameters for motif  $M$ and $H^M \in \mathbb{R}^{N \times F}$ is the output of the Conv. Unit.

\subsection{Combining Multiple Motifs}
\begin{figure}[t]
\includegraphics[width=\linewidth]{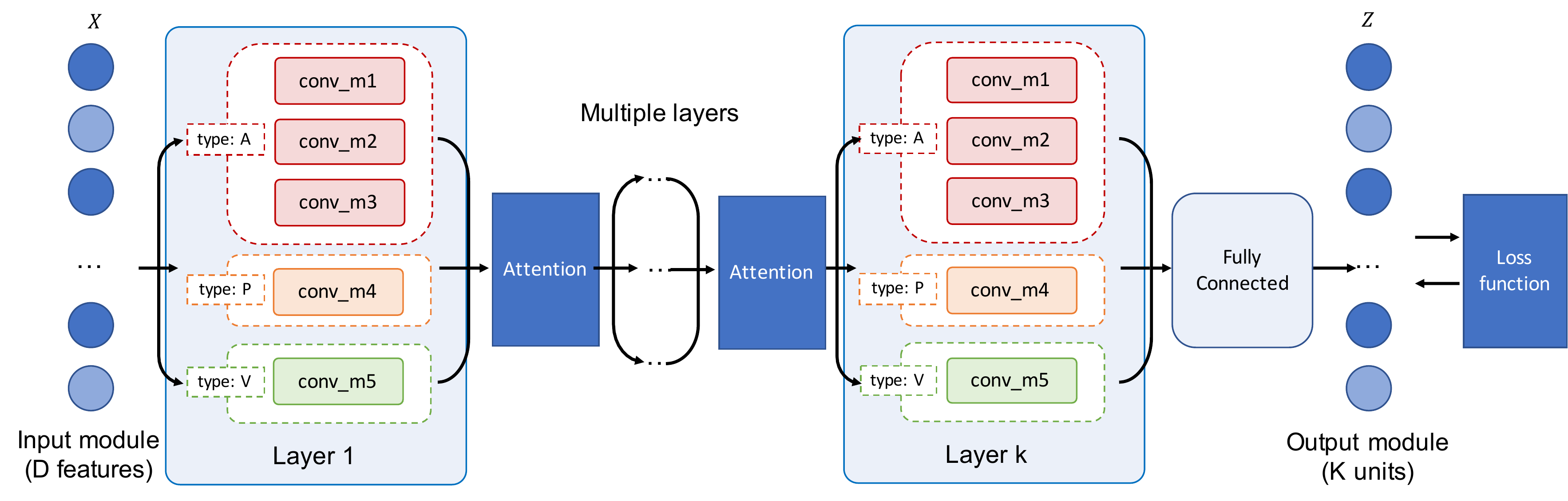}
\vspace{-18pt}
\caption{Architecture of \textit{Motif-CNN} for motifs in Fig.~\ref{fig:motif}}
\label{fig:arch}
\vspace{-13pt}
\end{figure}
\label{sec:arch}

So far, we have described a Conv. Unit that uses a single motif to obtain a first-order feature map.  
Since different motifs can vary in their importance for a task,  we face the challenge of appropriately weighting
the extracted features at the end of each layer for effective feature propagation. A direct solution is to use a set of $U$ weight parameters to combine the outputs of different motifs. However, this fails to capture varying levels of importance of motifs across nodes. We are inspired by recent advances in \textit{attention} mechanisms originally introduced for neural machine translation~\cite{attention} that allows various parts of the input to contribute differently while learning a combined representation. We propose a motif-attention model to dynamically weight the importance of different motifs for each node. \\
\textbf{Motif-Attention:}
We use the scaled dot-product form (introduced in ~\cite{vaswani2017attention}) for the attention model to compute the output at node $v_i$ given by:
\scriptsize
\begin{align}
h(v_i) = \sum\limits_{k=1}^U \alpha_{k,i} h^{k}(v_i)  \hspace{5pt}\alpha_{k,i} = softmax_k(e_{k,i}) = \frac{\exp(e_{k,i})}{\sum\limits_{j=1}^U \exp(e_{j,i}) } \label{eqn:attention}
\end{align}
\normalsize
where $e_{k,i} = a(h^{k} (v_i), z_k ) = \frac{z_k^T h^{k} (v_i)}{\sqrt{|z_k|}}$ are attention co-efficients that indicate the importance of motif $M_k$ to node $v_i$,
$z_k$ is a shared attention vector for motif $M_k$ describing the informativeness of $M_k$ across different nodes and $h^k(v_i)$ is the output of Conv. Unit for $M_k$ (Eqn.~\ref{eqn:conv_full}). 
The  architecture, illustrated in Fig.~\ref{fig:arch} for the motifs shown in Fig.~\ref{fig:motif}, comprises multiple stacked layers followed by a fully connected layer.
In each layer, the activations of all Conv. Units are combined via attention, to feed as input to the next layer, 
\textit{i.e.}, the output at layer $l$ denoted by $H_{l}$, is computed by Eqn.~\ref{eqn:attention} with $H_{0} = X$.

For a $K$-class classification setting, the output layer has $K$ units and applies the softmax activation function to obtain $Z \in \mathbb{R}^{N \times K}$. We use a cross-entropy loss function given by,
\scriptsize
\begin{equation}
 L =  - \sum\limits_{l \in Y_{{}_L}} \sum\limits_{k=1}^K Y_{lk} \log Z_{lk}
\end{equation}
\normalsize
For multi-label classification, we use $K$ sigmoid units in the output layer and apply the binary cross-entropy loss function.

\subsection{Complexity analysis}
We analyze the complexity of our model in two parts: \\
\textbf{Pre-Computation of $\mathcal{A}^M$}: The Motif-Adjacency Tensor, which is independent of the architecture is pre-computed for all motifs.
In this paper, we focus on motifs of upto 3 nodes. 
The cost of computing $\mathcal{A}^M$ for triangles is $O(|E|^{1.5})$~\cite{triangle}. For non-triangle 3-node motifs, each pair of neighbors can be examined for all nodes giving a complexity of $\Theta(\sum_{j}d_j^2)$ ($d_j$ is the degree of node $v_j$), with superior efficient algorithms in practice~\cite{mapreduce}.

For larger motifs, subgraph matching can be used with approximate sampling strategies for practical  efficiency.\\
\textbf{Model Training:}
The complexity of single layer is a function of the number of motifs $U$ (typically $<5$) and density of each $\mathcal{A}^M$, given by $ O(\sum_{i=1}^U |\mathcal{A}^{M_i}| D F)$.
In practice, the number of roles $K_M$ is at most $3$ and the role-specific matrices are sparser than the original adjacency matrix, giving an average-case complexity $ O(U |E| DF)$. Thus, we observe linear scaling with $U$ in comparison to GCN with $O(|E| DF)$.

\label{sec:impl}
An efficient implementation of Motif-CNN using sparse-dense matrix operations in Tensorflow~\cite{tensorflow} is publicly available~\footnote{https://github.com/aravindsankar28/Meta-GNN}.

\subsection{Discussion on Motif-based Convolution}
\label{sec:disc}
In this section, we demonstrate that our motif-based convolution (described in Eqn.~\ref{eqn:conv_partial}) can be expressed as \textit{standard} convolution over motif instances followed by mean pooling. 

Recall that a conventional CNN scans a \textit{square} filter to extract features through an inner product between the filter parameters and features. 
Similarly, we interpret the \textit{motif} filter as scanning the local neighborhood of a target node 
to compute an inner product over each instance. 
The output of the standard convolution operation at node $v_i$ and motif $M$, $h^M_s(v_i)$, at each instance $S_{v_i} = (V_S, E_S) \in I^M_{v_i}$ is given by:\\
\scriptsize
\begin{equation}
h^M_s(v_i) = w_0 x_i + \sum\limits_{v_j \in V_S - \{ v_i\}} w_{{}_{\phi_M (\psi_S (v_j))}} x_{j}  \; \; \; \forall \;  S_{v_i} \in I^M_{v_i}
\label{eqn:conv_eqn}
\end{equation}
\normalsize

Using a mean pooling operation to aggregate the outputs of convolution similar to traditional CNN architectures, we get:
\scriptsize
\begin{align}
h_{new}^M (v_i) &= \sigma \Bigg( \texttt{pool} \Big( \Big\{ h^M_s(v_i)  \; : \;  S_{v_i} \in I^M_{v_i}   \Big\}\Big) \Bigg) = \sigma \Bigg( \frac{1}{L_i} \sum\limits_{s=1}^{L_i} h^M_s (v_i) \Bigg) \nonumber \\  
&=   \sigma \Bigg(  w_0 x_i +  \frac{1}{L_i} \sum\limits_{s=1}^{L_i} \sum\limits_{v_j \in V_S - \{ v_i\}} w_{{}_{\phi_M (\psi_S (v_j))}} x_{j}  \Bigg) 
\label{eqn:pool_eqn}
\end{align}
\normalsize
where $h_{new}^M(v_i)$ is the output at node $v_i$ after pooling.
On careful inspection of Eqn.~\ref{eqn:pool_eqn}, we can express the second term as a sum over each node $v_j$ in $G$ weighted by the occurrence count of $v_j$ in the context of target $v_i$ in role $k$, thus reducing to Eqn.~\ref{eqn:conv_partial}. 
Thus, Eqn.~\ref{eqn:pool_eqn} provides an interpretation of Eqn.~\ref{eqn:conv_partial} as a \textit{standard convolution} over motif instances, followed by \textit{mean pooling}. This provides a strong basis for our motif-based formulation, since it generalizes primitives of conventional convolution and pooling operations to graphs. 
Note that in the trivial case of using an ``edge" as the only motif, our model (Eqn.~\ref{eqn:conv_full}) approximately reduces to GCN~\cite{iclr17}. Thus, \textit{Motif-CNN} generalizes state-of-the-art graph CNNs through high-order structures or motifs.

\section{Experiments}
\label{sec:expts}
\newcolumntype{K}[1]{>{\centering\arraybackslash}p{#1}}
\begin{table}[t] \scriptsize
\centering
\begin{tabular}{@{}p{0.2\linewidth}K{0.11\linewidth}K{0.11\linewidth}K{0.06\linewidth}K{0.15\linewidth}@{}}
\toprule
\textbf{Dataset} & \textbf{$|V|$} & \textbf{$|E|$} & \textbf{$D$} & Classes \\
\midrule
Flickr & \num[group-separator={,}]{13696} & \num[group-separator={,}]{1354461} & \num[group-separator={,}]{1000} & 10    \\
LinkedIn & \num[group-separator={,}]{7124} & \num[group-separator={,}]{39649} & \num[group-separator={,}]{2394} & 3  \\
\bottomrule
\end{tabular}
\vspace{-5pt}
\caption{Statistics of Flickr and LinkedIn social networks}
\label{homo_dataset_stats}
\vspace{-10pt}
\end{table}
In this section, we present experiments on homogeneous and heterogeneous graph datasets. We compare against three graph CNN methods a) DCNN \cite{dcnn} b) GCN \cite{iclr17} and c) Graph-CNN \cite{graph-cnn}. We exclude spectral methods as
they have been shown inferior to GCN. We also compare against a simpler model \textit{Motif-CNN-A} that uses a weighted combination of motifs instead of attention in each layer. We additionally compare against multiple state-of-the-art node classification techniques to present a comprehensive evaluation. 

\subsection{Homogeneous Graphs}
We conduct experiments on social network datasets with node attributes on semi-supervised node classification. 
We use two real-world social media datasets from Flickr~\cite{flickr} and LinkedIn~\cite{rase} in our experiments. In Flickr, the graph is described by the friendship network among users, node attributes by user interest tags and classes by user interest groups.
LinkedIn is a set of ego-networks with node attributes given by user profiles and classes by tags assigned by the ego-user to his friends into various categories, such as classmates, colleagues, etc. We only include ego-networks with at least 10\% labels per class. Flickr corresponds to a multi-label scenario, while LinkedIn is multi-class.
Table.~\ref{homo_dataset_stats} illustrates the statistics of the two datasets. \\
\textbf{Experimental setup:}
In each dataset, we randomly sample 20\% of the labeled examples for training, 10\% for validation and the rest for testing. We repeat this process 10 times, and report the average performance in terms of both Micro-$F_1$ and Macro-$F_1$.
Unless otherwise stated, we train a 3-layer \textit{Motif-CNN} with ReLU activations and tune hyper-parameters (learning rate, dropout rate and number of filters per motif) based on the validation set. We train all graph CNNs for a maximum of 200 epochs using Adam~\cite{adam} with windowed early stopping on the validation set.
Since we expect motifs such as triangles ($\Delta$) to be discriminative in social networks, we experiment with \textit{$\Delta$} and \textit{($\Delta$ + edge)}.
Additionally, we compare against two standard baselines for SSL:  ICA~\cite{collective} and Planetoid~\cite{planetoid}.\\
\textbf{Experimental results:}
We summarize the classification results in Table.~\ref{homo_class_results}. \textit{Motif-CNN ($\Delta$ + edge)} outperforms other benchmark algorithms and achieves gains of 6\% in Flickr and 11\% in LinkedIn (Macro-$F_1$) over the next-best method.
\textit{Motif-CNN ($\Delta$)} suffers from sparsity for nodes of low degree, which is offset by \textit{Motif-CNN ($\Delta$+edge)} which uses both edge and triangle patterns.
This highlights the ability of \textit{Motif-CNN} in learning feature associations through triangle patterns that are important in social networks. \textit{Motif-CNN} outperforms the basic model \textit{Motif-CNN-A}, justifying the choice of dynamically weighting the features from multiple motifs. 

\begin{table}[t] \scriptsize
\begin{tabular}{@{}p{0.35\linewidth}K{0.13\linewidth}K{0.13\linewidth}K{0.13\linewidth}K{0.13\linewidth}@{}}
\toprule
\multirow{2}{*}{\textbf{Method}} & \multicolumn{2}{c}{\textbf{Flickr}}  &   \multicolumn{2}{c}{\textbf{LinkedIn} } \\
\cmidrule(lr){2-3} \cmidrule(lr){4-5}
 & \textbf{Micro-F1} & \textbf{Macro-F1} & \textbf{Micro-F1} & \textbf{Macro-F1}\\
 \midrule
DCNN &46.39  & 46.96 & 70.27 & 54.81 \\
GCN & 41.71 & 42.24 & 71.91 & 57.53 \\
Graph-CNN & 40.61 & 41.86 & 70.02 & 53.57 \\
Planetoid & 31.53 & 31.39 & 62.49 & 43.63 \\
ICA & 28.53 & 22.45 & 63.39 & 47.55 \\
 \midrule
Motif-CNN ($\Delta$) & 47.05 & 47.45 & 72.13 & 58.75 \\
Motif-CNN-A ($\Delta$+ edge) & 46.29 & 46.94  & 74.42 & 63.51\\
Motif-CNN ($\Delta$+ edge) & \textbf{49.10} & \textbf{49.56}& \textbf{74.75} & \textbf{63.57} \\
\bottomrule
\end{tabular}
\vspace{-8pt}
\caption{Classification results on Flickr and LinkedIn.}
\label{homo_class_results}
\vspace{-10pt}
\end{table}
\subsection{Heterogeneous Graphs}
\newcommand{\Arrow}[1]{%
\parbox{#1}{\tikz{\draw[->](0,0)--(#1,0);}}
}
We conduct classification experiments on heterogeneous graphs using three real-world datasets (statistics in Table.~\ref{hin_dataset_stats}):
 \textbf{DBLP-A}: This is a bibliographic citation graph composed of 3 node types: author ($A$), paper ($P$) and venue ($V$), connected by three link types: $P \Arrow{0.2cm} P$, $A\mbox{-}P$ and $P\mbox{-}V$. We use a subset of DBLP~\cite{pathsim} with text features of papers to classify authors based on their research areas.  \\
 \textbf{DBLP-P}: This dataset has the same schema as DBLP, but the task is to classify research papers. The categories of papers are extracted from Cora~\cite{cora}. \\
 \textbf{Movie}: We use MovieLens~\cite{movielens} to create a graph with 4 node types: movie ($M$), user ($U$), actor ($A$) and tag ($T$) linked by 4 types: $U\mbox{-}M$, $A\mbox{-}M$, $U\mbox{-}T$ and $M\mbox{-}T$, with features available for actors and movies, for movie genre prediction which is multi-label classification. 
\begin{table}[h] \scriptsize
\vspace{-5pt}
\centering
\begin{tabular}{@{}p{0.2\linewidth}K{0.11\linewidth}K{0.11\linewidth}K{0.06\linewidth}K{0.15\linewidth}@{}}
\toprule
\textbf{Dataset} & \textbf{$|V|$} & \textbf{$|E|$} & \textbf{$|\mathcal{L}|$} 
& Classes \\
\midrule
DBLP-A & \num[group-separator={,}]{11170} & \num[group-separator={,}]{24846} & 3  & 4  \\
DBLP-P & \num[group-separator={,}]{35770} & \num[group-separator={,}]{131636} & 3 & 10  \\
Movie &  \num[group-separator={,}]{10441} & \num[group-separator={,}]{99509} & 4 & 6  \\
\bottomrule
\end{tabular}
\vspace{-8pt}
\caption{Statistics of heterogeneous graph datasets}
\label{hin_dataset_stats}
\vspace{-10pt}
\end{table}\\
\textbf{Experimental setup:}
We sample 10\% of the labeled examples for training, 10\% for validation and rest for testing. For node types that do not have features, we assume 1-hot encoded inputs for the graph convolutional models. For \textit{Motif-CNN}, we use all \textit{relevant} 3-node motifs that indicate semantic closeness based on the graph schema.
We provide the details of these motifs online\footnote{https://sites.google.com/site/motifcnn/}.
We also present comparisons against multiple heterogeneous SSL baselines given below: \\
 \textbf{LP-Metapath}: SSL algorithm that utilizes metapath-specific Laplacians to jointly propagate labels and learn weights for different metapaths~\cite{lp_metapath}.\\
 \textbf{LP-Metagraph}: SSL algorithm based on  an ensemble of metagraph guided random walks~\cite{metaijcai17}. \\
 \textbf{Column Network (CLN)}:  Deep neural network for classification in multi-relational graphs~\cite{column}.\\
 \textbf{metapath2vec}: Heterogeneous network embedding model that uses metapath-based random walks~\cite{m2v}. 

For metapath-based methods, we provide all relevant metapaths and for LP-Metagraph, we use all metagraphs with $\leq 4$ nodes and report the best performance among their three ensemble methods. Note that LP-Metapath and LP-Metagraph are not applicable for multi-label classification.
\begin{table}[t] \scriptsize
\begin{tabular}{@{}p{0.18\linewidth}K{0.09\linewidth}K{0.096\linewidth}K{0.09\linewidth}K{0.096\linewidth}K{0.09\linewidth}K{0.096\linewidth}@{}}
\toprule
\multirow{2}{*}{\textbf{Method}} & \multicolumn{2}{c}{\textbf{DBLP-A}}  &   \multicolumn{2}{c}{\textbf{DBLP-P} } & \multicolumn{2}{c}{\textbf{Movie} } \\
\cmidrule(lr){2-3} \cmidrule(lr){4-5} \cmidrule(lr){6-7}
 & \textbf{Mic-F1} & \textbf{Mac-F1} & \textbf{Mic-F1} & \textbf{Mac-F1} &\textbf{Mic-F1} & \textbf{Mac-F1} \\
\midrule
DCNN & 69.68 & 69.20 & x & x & 49.91 & 46.72\\
GCN  & 81.34 & 81.29 & 71.30 & 53.14 & 55.67 & 53.85\\
Graph-CNN & 72.89 & 73.04 & 65.58 & 52.85 & 52.07 & 49.53 \\
LP-Metapath & 82.77 & 82.86 & 62.15 & 52.41 & - &- \\
LP-Metagraph & 83.03 & 83.11 & 62.97 & 57.08 & - & -  \\
CLN & 80.99 & 80.94 & 66.71 & 52.19 & 57.11 & 47.71\\
metapath2vec & 83.37 & 83.43 & 70.10 & 61.89 & 60.00 &  60.00 \\
\midrule
Motif-CNN-A & {86.44} & {86.49} & {71.19} & {57.05}  & {62.41} & {60.64}\\
Motif-CNN & \textbf{86.71} & \textbf{86.78} & \textbf{74.76} & \textbf{64.17}  & \textbf{62.61} & \textbf{61.19}\\
\bottomrule
\end{tabular}
\caption{Classification results on DBLP-A, DBLP-P and Movie, \\
  - indicates \textit{not applicable}, x indicates \textit{does not scale}.}
\label{hin_results}
\vspace{-15pt}
\end{table}\\
\textbf{Experimental results:}
From Table.~\ref{hin_results}, we observe that \textit{Motif-CNN} achieves gains of 7\% and 21\% (Macro-$F_1$) over other graph CNN models while gaining 4\% and 4\% overall in DBLP-A and DBLP-P respectively. Since the task of classifying research papers in DBLP-P is more fine-grained, the attention mechanism significantly improves performance by appropriately weighting the importance of different motifs.
In DBLP-P, DCNN does not scale due to $O(N^2)$ space complexity with the authors' implementation. In Movie, \textit{Motif-CNN} performs the best
with gain of 14\% Macro-$F_1$ and 12\% Micro-$F_1$ over graph CNN models.
Thus, \textit{Motif-CNN} convincingly outperforms all graph CNN models on heterogeneous datasets, while gaining over other state-of-the-art classification methods by a smaller margin. 
The relatively higher gains in heterogeneous datasets shows the power of capturing high-order features through relevant patterns.

\subsection{Computational Efficiency}
\begin{figure}[t]
\vspace{-5pt}
\subfigure{
        \centering
        \includegraphics[width=0.45\linewidth]{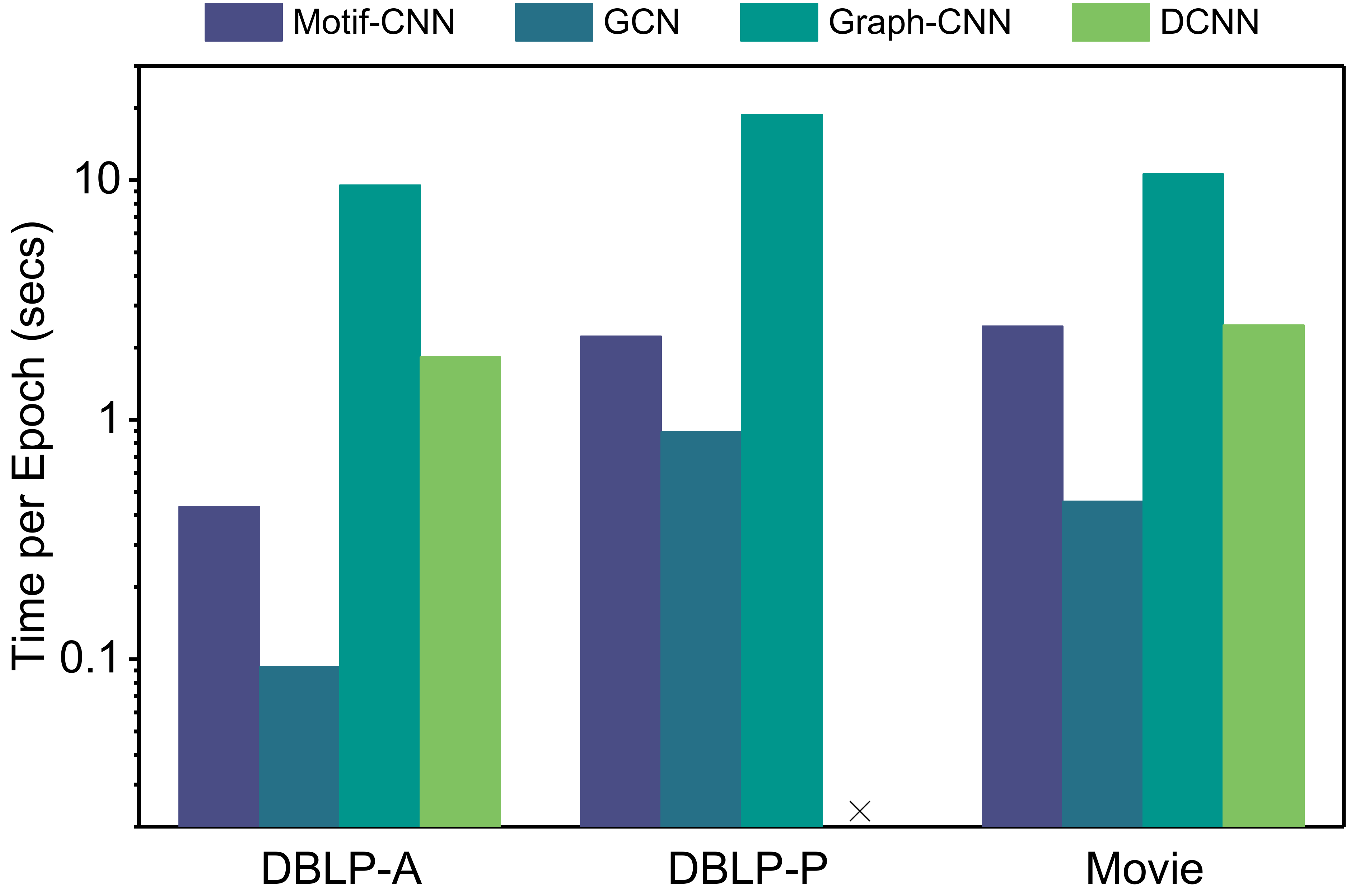}
        \label{fig:time_epoch}}
\subfigure{
        \centering
        \includegraphics[width=0.46\linewidth]{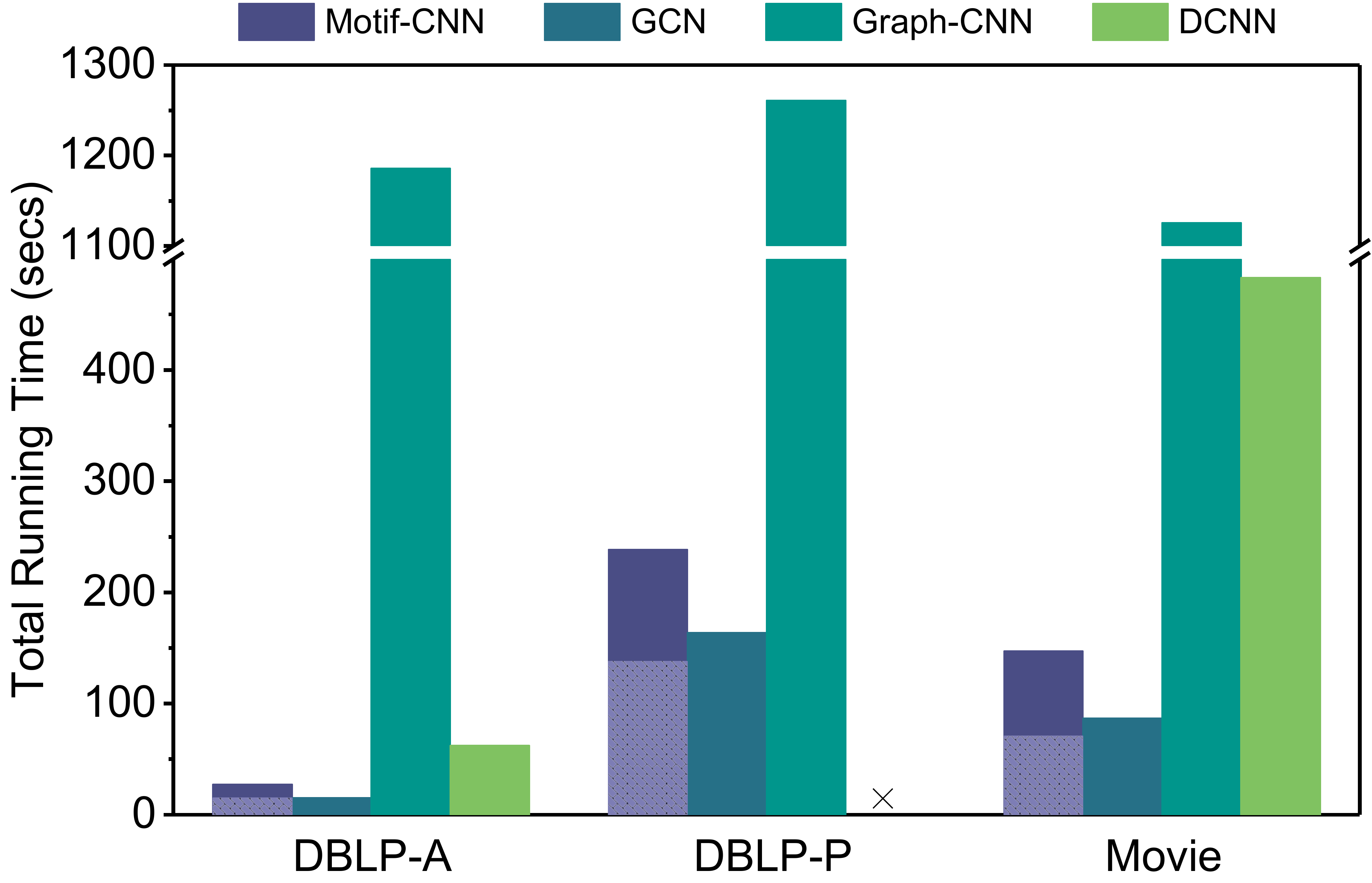}
        \label{fig:time_total}}
\vspace{-10pt}
\caption{Comparison of (a) Time per epoch and (b) Total runtime with the shaded area denoting pre-computation time}
\vspace{-5pt}
\label{fig:runtime}
\end{figure}

We report running times on an Intel(R) Xeon(R) CPU E5-2699 v4 2.20 GHz system with 8 cores and 64 GB memory. \\
\textbf{Computation Time:} 
We compare the model training time per epoch of \textit{Motif-CNN} versus other graph CNNs on the heterogeneous graph datasets in Fig.~\ref{fig:runtime}. We find that \textit{Motif-CNN} is quite efficient in practice and comes second only to GCN (which is expected - see Sec.~\ref{sec:disc}) while Graph-CNN is orders of magnitude slower since it entails dense matrix operations.

\begin{figure}
\vspace{-1pt}
\includegraphics[width=\linewidth]{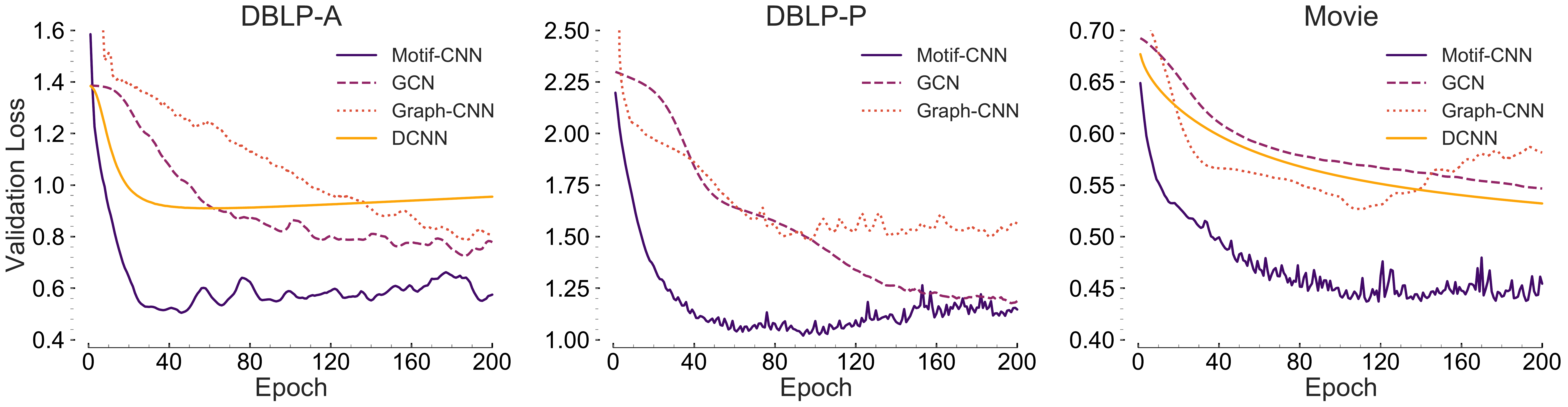}
\vspace{-15pt}
\caption{Plots of validation loss w.r.t. epochs for all Graph CNNs}
\label{fig:convergence}
\vspace{-15pt}
\end{figure}

We evaluate the efficiency of different models by comparing the total running time till convergence.
Although the pre-computation cost is a noticeable (but not substantial) portion of the total time, \textit{Motif-CNN} is reasonably close to GCN as its rapid convergence trades off the cost of pre-computation.\\
\textbf{Convergence:}
We compare the convergence rates of different graph CNNs by depicting the validation set loss
in Fig.~\ref{fig:convergence} on using the same model configuration and hyper-parameters. Overall, \textit{Motif-CNN} achieves lower error and faster convergence in comparison to other graph CNNs since it leverages multiple relevant patterns simultaneously.

\captionsetup[subfigure]{labelformat=empty}

\section{Related work}
\label{sec:related}
We organize related work in three sections: a) Graph CNNs b) Graph Motifs and c) Semi-supervised classification. \\
\textbf{Graph CNNs} can be categorized in two general directions:

\textit{Spectral:} Convolution is defined via the Fourier Transform described by eigenvectors of the Graph Laplacian \cite{iclr14,nips16}. As noted in Sec.~\ref{sec:intro}, spectral CNNs cannot be transferred across different graphs. 

\textit{Spatial:}
These methods directly generalize convolution in the graph domain through immediate neighborhood proximity~\cite{dcnn,icml16,graph-cnn,iclr17}. 
As discussed in Sec.~\ref{sec:intro}, all these methods 
are limited to learning just a single type of filter through convolution, rendering them incapable of  
modeling semantic relevance in heterogeneous graphs. \\
\textbf{Graph Motifs:}
Motifs are high-order structures that are
crucial in many domains such as neuroscience \cite{motif_neuro}, bioinformatics~\cite{motif_bio,reactionminer} and social networks. Recent work has explored motifs in clustering~\cite{science}, clique detection~\cite{mclique},
strong tie detection~\cite{strong_tie}, graph classification~\cite{graphlet} and ranking~\cite{ranking}. 
In contrast, we employ motifs to define the receptive field around a target node of interest for graph convolution. \\
\textbf{Semi-supervised node classification:} In this paper, we focus on graph-based SSL which has been well-studied for homogeneous graphs and is often termed Collective Classification~\cite{collective} with node features.

In heterogeneous graphs, metapaths~\cite{lp_metapath} and metagraphs~\cite{metagraph} have been leveraged to develop skip-gram~\cite{m2v} and deep learning models~\cite{column} for SSL~\cite{metaijcai17}.
All these techniques use metapaths or metagraphs to model just the similarity between a pair of nodes of the same type but cannot utilize the features of all node types for classification. In contrast, we propose a unified framework to extract local features through more general high-order structural patterns.

\section{Conclusion and Future Work}
In this paper, we have introduced a novel convolution operation that uses motifs to capture the key aspects of local connectivity and translation invariance in graphs.
We proposed \textit{Motif-CNN} that effectively fuses information from multiple patterns to learn high-order features through deeper layers. Our experiments demonstrate significant gains over existing graph CNNs especially on heterogeneous graphs.

We identify multiple interesting directions for future work. To scale the model to large graphs, 
sampling methods~\cite{graphsage} can be explored to approximate motif-based neighborhoods. 
Further avenues for future work include extensions of our framework to temporally evolving graphs~\cite{dysat} and user behavior modeling for recommender systems~\cite{latent,cikm18}.

\bibliographystyle{named}
\scriptsize
\bibliography{ijcai18}

\end{document}